\documentclass[a4paper,fleqn]{cas-dc}

\usepackage{bm}
\usepackage{amssymb}
\usepackage[authoryear]{natbib}
\usepackage{float}
\usepackage{supertabular}
\usepackage{algorithm}
\usepackage[noend]{algpseudocode}
\usepackage{amsmath}

\DeclareMathOperator*{\argmax}{arg\,max}
\DeclareMathOperator*{\argmin}{arg\,min}

\def\tsc#1{\csdef{#1}{\textsc{\lowercase{#1}}\xspace}}
\tsc{WGM}
\tsc{QE}
\tsc{EP}
\tsc{PMS}
\tsc{BEC}
\tsc{DE}

\begin{document}
\let\WriteBookmarks\relax
\def\floatpagepagefraction{1}
\def\textpagefraction{.001}
\shorttitle{}
\shortauthors{Adnan Haider et~al.}

\title[mode = title]{A Distributed Optimisation Framework Combining  Natural Gradient with Hessian-Free for Discriminative Sequence Training} 

\author{Adnan Haider}
\fnmark[1,2]
\ead{adnan_haider@apple.com}
\fntext[fn1]{Work was done while Adnan Haider was at Cambridge University}

\address{Cambridge University Engineering Department, Trumpington Street, Cambridge, CB2 1PZ U.K.}

\author{Chao Zhang}
\fnmark[2]
\ead{cz277@cam.ac.uk}

\author{Florian L. Kreyssig}
\fnmark[3]
\ead{flk24@cam.ac.uk}

\author{Philip C. Woodland}
\cormark[1]
\ead{pcw@eng.cam.ac.uk}

\fntext[fn2]{Equal contributions.}
\fntext[fn3]{Funded by an EPSRC Doctoral Training Partnership Award.}
\cortext[cor1]{Corresponding author.}

\begin{abstract}
This paper presents a novel natural gradient and Hessian-free (NGHF) optimisation framework for neural network training that can operate efficently in a distributed manner. It relies on the linear conjugate gradient (CG) algorithm to combine the natural gradient (NG) method with local curvature information from Hessian-free (HF) or other second-order methods. 
A solution to a numerical issue in CG allows effective parameter updates to be generated with far fewer CG iterations than usually used (\textit{e.g.} 5-8 instead of 200). This work also presents a novel preconditioning approach to improve the progress made by individual CG iterations for models with shared parameters. Although applicable to other training losses and model structures, NGHF is investigated in this paper for lattice-based discriminative sequence training for hybrid hidden Markov model acoustic models using a standard recurrent neural network, long short-term memory, and time delay neural network models for output probability calculation. 
Automatic speech recognition experiments are reported on  the multi-genre broadcast data set for a range of different acoustic model types. These experiements show that NGHF achieves larger word error rate reductions than standard stochastic gradient descent or Adam, while requiring orders of magnitude fewer parameter updates.

\end{abstract}


\maketitle
\emergencystretch 3em
\sloppy
\section{Introduction}
With the availability of increased computing power and appropriate parameter initialisation methods \citep{Hinton:2006b,hinton2012,glorot}, the hybrid automatic speech recognition (ASR) approach, {\it i.e.} the use of hidden Markov model (HMM) acoustic model with neural networks (NNs) for observation probability calculation, has achieved state-of-the-art performance on large vocabulary continuous speech recognition (LVCSR) tasks \citep{Xiong:2016a,Saon:2016a}. Whilst NNs can model the underlying non-linear manifold of speech data well \citep{hinton2012}, their deep structures create complex dependencies among parameters that can make these models difficult to train with standard, or variants of, stochastic gradient descent (SGD) \citep{Desjardins2015,grosse2015scaling}.

Due to the sequential nature of speech,
sequence-level training objectives are often used for ASR. It has been observed that training acoustic models using a discriminative loss function that not only maximises the probability of the reference transcription or recognition accuracy  but also minimises that of all competing hypotheses often improves ASR performance by a significant margin \citep{Lattice2,Povey2005,Graves:2014e2e,LFMMI}. Therefore discriminative loss functions have become a standard choice for state-of-the-art ASR systems \citep{Saon:2016a,Xiong:2016a,chiu2018StateoftheartSpeechRecognition,luscher2019RWTHASRSystems}. For training NNs with such a loss, SGD with training data shuffled at the utterance level is widely used. Often this use of SGD requires considerable skill in choosing suitable values for hyper-parameters such as the learning rate or the momentum coefficient,  
while second-order optimisation methods require fewer hyper-parameters and can lead to more effective parameter updates compared to SGD \citep{Xu2018}. These approaches leverage the local curvature information contained in the Hessian matrix ({\em i.e.} the second-order derivatives w.r.t. the model parameters) to overcome the issues that SGD encounters with highly non-linear and ill-conditioned loss functions \citep{becker1988improving}. However, a major weakness of second-order methods is that they are computationally inefficient for models with a large number of parameters. 
This is because scaling the gradient directions by the inverse of the Hessian matrix requires computation on the order of $\mathbb{O}(D^3)$ where $D$ is the number of model parameters.

A large batch second-order approach, the Hessian-free (HF) method, was proposed by \citet{Martens2010}. Instead of computing the Hessian matrix explicitly, HF finds an approximate solution by solving an equivalent system 
using the 
conjugate gradient (CG) algorithm. 
Compared to SGD with mini-batches typically containing only a few utterances, 
HF can yield more effective updates based on very large batches (\textit{e.g.} thousands of utterances
) and requires far fewer parameter updates to converge. 
Hence, it is easier to parallelise HF than SGD by distributing the training data in each large batch among all available computation units and accumulating the outcomes to yield a parameter update. 
However, in practice, HF 
can require more computation than SGD, since it often needs a large number (\textit{e.g.} 200) of CG iterations
for each parameter update
\citep{Martens2010,kingsbury2012,wiesler2013,Martens2015}.

Recently, the natural gradient (NG) method, which was originally proposed for training small models with maximum likelihood (ML) \citep{Amari1997}, has received renewed popularity for training deep NN models \citep{Pascanu2013a,Desjardins2015}. 
NG computes the steepest descent direction in the space of the model output distributions instead of the space of the model parameters,  
and can result in fewer generalisation errors and faster convergence \citep{Roux,bernacchia2018exact}.
NG requires calculating the product of the inverse of the  Fisher information (FI) matrix with the gradient estimate. As for second-order approaches, directly calculating the inverse of the FI matrix
has a $\mathbb{O}(D^3)$ computational complexity,
rendering it unusable for large models.
To reduce this cost, approaches such as enforcing a block diagonal structure on the FI matrix and approximating the diagonal blocks as the Kronecker product of two smaller matrices have been introduced \citep{povey2014parallel,Martens2015,grosse2015scaling,George2018}.

In this paper, we propose an approach to stabilise the computation of the directional derivative that alleviates the numerical instability that can accompany CG when applied to NNs. To improve training on models with shared architectures, this work proposes a novel a preconditioning approach to speed up the progress made by CG.
Furthermore, in order to estimate the inverse of the FI matrix without making any assumptions about its structure, we propose applying CG to NG for discriminative sequence training.
As a result, CG can provide a common optimisation framework to use NG jointly with HF or other second-order approaches. We, therefore, propose the NGHF method that combines the advantages of NG and HF and further improves the stability and performance of NN model training.
The efficacy of the proposed optimisation framework and the NGHF method is evaluated by conducting ASR experiments on a 200 hour multi-genre broadcast (MGB) speech recognition dataset \citep{Bell2015,Woodland2015}. Discriminative sequence training with hybrid HMM acoustic models is studied, which use long short-term memory networks (LSTMs) \citep{Hochreiter1997},  recurrent neural networks (RNNs) \citep{rumelhart,elmanRNN}, or time-delayed neural networks (TDNNs) \citep{timedelayWaibel1,DanTDNN} for output probability calculations. 
Compared to SGD and Adam \citep{kingma2014adam}, NGHF is demonstrated to have a faster convergence speed and can result in lower word error rates (WERs). 

This paper is organised as follows. Section~\ref{sec:relatedworks} and \ref{sec:HFOpt} review related work on training NNs and the HF method with CG. Section~\ref{sec:ImprovingHF} presents our improved implementation of CG along with the distributed optimisation framework.
Sections~\ref{sec:NaturalGrad} and \ref{sec:NGHF} describe
the proposed CG-based NG and NGHF methods for discriminative sequence training. The experimental setup and results are given in Sec.~\ref{sec:expsetup} and Sec.~\ref{sec:ExpResults}, followed by a discussion in Sec.~\ref{sec:Discussion} and finally conclusions.

Compared to our previous papers \citep{haider2017,haider2018}, this paper proposes a more general approach to regulate the scaling and rotation induced on the gradient direction during NG descent learning with second-order methods (Sec.~\ref{sec:NGHF}). All techniques are described more completely and comprehensively. 
The design of the distributed CG-based optimisation method is fully presented (Sec. \ref{ssec:gbdist}). The description and analysis of how to improve the numerical stability of CG (Sec. \ref{ssec:cgstab}) and a preconditioning method that improves the performance for models with shared parameters (Sec. \ref{sec:CGpredcond}) are included.
The experimental results have been extended and now include comparisons of the proposed method for TDNNs, LSTMs and other recurrent models.

\section{Related work}
\label{sec:relatedworks}
Starting from layer-by-layer pre-training and improved random initialisation methods \citep{Hinton2006,glorot},
optimisation improvements have played a critical role in the development of deep learning. Various methods have been applied to improve the stability and efficiency for training complex NN structures with SGD. Even though the use of these methods has significantly improved NN training, the task of finding a more suitable optimisation procedure for supervised NN training continues to be an active area of research. This section first reviews different SGD, HF, and NG methods, with a focus on distributed computation. Then discriminative sequence training is reviewed, which is a key focus of this paper.

\subsection{Synchronous and Asynchronous SGD}
Adapting the learning rates during training is known to improve the stability and convergence speed of SGD. Rule-based strategies can be used to decay the learning rate shared by all DNN parameters \citep{Renals1992,Bottou2010,Senior:2013a}. Methods such as RMSProp \citep{Tieleman2012}, AdaGrad \citep{AdaGrad}, AdaDelta \citep{zeiler2012adadelta}, 
and Adam \citep{kingma2014adam} assign a separate learning rate to each model parameter. These are updated based on previous gradients. 
Parameter updates can also be derived as the output of an RNN based on the gradients \citep{Andrychowicz2016}.
The methods developed in this paper do not require tuning of learning rates and implicitly assign separate learning rates to each parameter. 

In standard SGD, greater parallelisation is achieved by using a larger mini-batch. 
The (equivalent) mini-batch size can be increased in a synchronous manner, which requires every ``worker'' process to calculate the gradients for their mini-batch for the same copy of the model parameters. These are then combined \citep{zinkevich2010parallelized}. 
The main issue of synchronous distributed SGD is the high transmission cost \citep{Seide2014,Strom2015}. 
Asynchronous SGD (ASGD) can reduce the transmission cost by allowing each worker to keep a distinct copy of the model and to communicate independently with a centralised parameter server \citep{Dean2012,Heigold2014} or other workers \citep{zhangw2019a,zhangw2019b}. This also increases the number of updates.
The blockwise model update filtering method divides workers into blocks and averages the models produced by each block with a mechanism analogous to the use of momentum in standard SGD \citep{chen2016scalable,Ladkat2019}. The optimisation methods developed in this paper can be used in a highly parallel and distributed manner, 
which relies on the use of very large batches and a small number of updates to efficiently and accurately exploit parallelisation.

\subsection{HF and NG based Approaches}
\label{ssec:reviewnghf}
The HF optimisation framework was initially proposed to minimise the squared errors for auto-encoders based on the linear CG algorithm \citep{shewchuk1994introduction} and the Gauss-Newton (GN) approximation \citep{Martens2010}. Later studies applied the approach to cross-entropy (CE) training for small scale ASR and hand-written digit classification \citep{Vinyals2012,wiesler2013}. \citet{kingsbury2012} extended HF to lattice-based discriminative sequence training for LVCSR with a large batch size, which allows the data to be processed in parallel based on a master/worker structure. The workers compute the gradients and the curvature information based on the same copy of the model parameters in a distributed fashion, while the master process collects the outputs from all workers to perform the CG algorithm to update the model parameters. Momentum and pre-conditioning can be integrated into this optimisation framework as shown in later studies \citep{Sainath2013b,Sainath2013a}. 
To reduce the computation workload of the master, only a small percentage (\textit{e.g.} 1\%) of the training data is sampled for use in each CG iteration \citep{Martens2010,kingsbury2012}.

Compared to the original work on NG that is aimed at training small models with the ML criterion \citep{Amari1997}, more recent work focuses on training large NN models with tens of millions of parameters, in which case it is computationally impractical to estimate the inverse of the FI matrix. 
To overcome this issue, most studies assume that the FI matrix has some form of block diagonal structure \citep{Roux, povey2014parallel,Martens2015,grosse2015scaling,George2018}. 
Kronecker-factored approximate curvature (K-FAC) is such an approach that assumes the parameters from different layers are independent and approximates each block of the FI matrix as the Kronecker product of two smaller matrices \citep{Martens2015}. 
\citet{povey2014parallel} proposed a similar idea and applied it to
lattice-free maximum mutual information training \citep{LFMMI}, 
which was observed to yield no performance loss when simply averaging the models produced by the different workers configured to operate asynchronously.
Interestingly, the second-order momentum used in Adam can be viewed as an approximate diagonal FI matrix that assumes every model parameter is independently estimated \citep{kingma2014adam}. 
\citet{Schulman2015} proposed using CG to estimate the inverse of the FI matrix without enforcing any assumptions for reinforcement learning. \citet{haider2017} proposed a similar method for discriminative sequence training for ASR.

\subsection{Discriminative Sequence Training}
\label{sec:discriminative}
An ASR system aims to convert the acoustic feature sequence $\mathbf{O}$ of a speech utterance to its underlying word sequence $\mathbf{W}^{\text{ref}}$. In practice this is often achieved by finding the hypothesised word sequence $\hat{\mathbf{W}}$ that gives the maximum posterior probability $P(\mathbf{W} |\mathbf{O})$, \textit{i.e.}   
 \begin{align}
\label{equ1}
\hat{\mathbf{W}}&=\argmax\nolimits_{\mathbf{W}} P(\mathbf{W} |\mathbf{O}).
\end{align}
In the noisy source-channel framework for ASR, $P(\mathbf{W} |\mathbf{O})$ is calculated using
 an HMM-based
 acoustic model that produces $p(\mathbf{O}|\mathbf{W},\bm{\theta})$ (denoted as $p_{\bm{\theta}}(\mathbf{O}|\mathbf{W})$ in this paper) and a language model (LM) that estimates $P(\mathbf{W})$.
 In practice, the LM is first trained separately and then combined at test-time to decode the input acoustic feature vectors for an utterance into the most probable word sequence.  

Let  $\mathcal{M}$ denote the space of all probability distributions $P_{\bm{\theta}}(\mathbf{W}|\mathbf{O})$ that can result from different parameter configurations of a chosen NN when employed within an HMM.  The goal of learning is to identify a viable candidate in $\mathcal{M}$  that achieves the greatest reduction in the empirical loss, which refers to 
the average loss over all training samples {w.r.t.} a given risk function while generalising well to new examples. 
To effectively train acoustic models, two forms of  discriminative sequence level criterion are commonly used.  The first form corresponds to the maximum mutual information (MMI) loss. For a single observation feature sequence $\mathbf{O}$, the MMI loss corresponds to
 \begin{align}
\label{eq:MMIloss}
\mathcal{L}_{\rm{MMI}}(\bm{\theta})=-\ln\dfrac{p_{\bm{\theta}}(\mathbf{O}|\mathbf{W}^{\text{ref}})^{\kappa}P(\mathbf{W}^{\text{ref}})}{\sum_{\mathbf{W}}p_{\bm{\theta}}(\mathbf{O}|\mathbf{W})^{\kappa}P(\mathbf{W})},
\end{align}
The method was initially proposed for spoken digit recognition \citep{bahl1986maximum}, 
where $\kappa$ is the acoustic scaling factor that re-scales the acoustic model  and language model scores to be in the same range \citep{MMI2}.  
The MMI loss can be viewed as  maximising the probability of $\mathbf{W}^{\text{ref}}$ while also minimising that of every competing hypothesis $\mathbf{W}$, and is thus a discriminative sequence training loss.  
When applied to LVCSR \citep{Lattice2}, the denominator of Eqn.~\eqref{eq:MMIloss} needs to be calculated efficiently and often relies on using word lattices for each utterance as a compact representation of all of the important competing hypotheses. 

The minimum Bayes risk (MBR) loss is another commonly used loss function for discriminative sequence training \citep{GOEL,MBR1}, which directly minimises ASR errors by using 
WER related metrics as the risk function. The MBR loss for a single observation feature sequence $\mathbf{O}$ is defined as
\begin{align}
\label{eq:MBRloss}
\mathcal{L}_{\text{MBR}}(\bm{\theta})= \dfrac{\sum_{\mathbf{W}}p_{\bm{\theta}}(\mathbf{O}|\mathbf{W})^{\kappa}P(\mathbf{W})\mathcal{A}(\mathbf{W},\mathbf{W}^{\text{ref}})}{\sum_{\mathbf{W}}p_{\bm{\theta}}(\mathbf{O}|\mathbf{W})^{\kappa}P(\mathbf{W})},
\end{align}
where $\mathcal{A}(\mathbf{W},\mathbf{W}^{\text{ref}})$ is the risk function measuring the difference between the reference and a competing hypothesis. For ASR, approximations to the number of word-level, phone-level, and HMM state-level errors are widely used as the risk function \citep{MPE,gibson2006hypothesis,MattShannon}.
When phone-level errors are used, the MBR loss is called minimum phone error (MPE) \citep{MPE}, which will be the MBR loss used in the experiments in this paper.
As for MMI, for LVCSR  word lattices of each training utterance are required since the calculation of the MBR loss involves all competing hypotheses. 

Lattice-based discriminative sequence training with an MMI or MBR loss has also been widely used to finetune NN-HMM hybrid systems \citep{Valtchev1995,kingsbury2009lattice,vesely,Su2013,Wiesler2015,zhang:2015HTK}, which are often initialised by frame-level training with the CE loss \citep{hinton2012}.  Lattice-free MMI uses a general 
phone-level recognition network to replace utterance-specific lattices, which enables efficient processing of multiple training utterances in parallel \citep{LFMMI}. 

\section{Hessian-free Optimisation}
\label{sec:HFOpt}
This section provides an overview of the HF optimisation framework \citep{Martens2010,kingsbury2012}. At the core of all first and second-order optimisation methods is Taylor's theorem.  Assuming the loss function $\mathcal{L}(\bm{\theta})$ is sufficiently smooth, the \textit{second-order Taylor approximation} employs the following quadratic function to locally approximate the  function as 
\begin{align}
\label{eq:2ndtaylor}
\mathcal{L}(\bm{\theta} +\Delta\bm{\theta}) \approx\mathcal{L}(\bm{\theta}) +  \Delta \bm{\theta}^{\text T}\nabla_{\bm{\theta}}\mathcal{L}(\bm{\theta})+ \frac{1}{2} \Delta \bm{\theta}^{\text T} \mathbf{H}  \Delta \bm{\theta}, 
\end{align}
where $\nabla_{\bm{\theta}}$ is the gradient operator in the space of $\bm{\theta}$, $\Delta\bm{\theta}$ represents an offset within a convex neighbourhood of $\bm{\theta}$, and $\mathbf{H}$ is the Hessian matrix of $\mathcal{L}$ w.r.t. $\bm{\theta}$, \textit{i.e.}  $\mathbf{H}=\nabla^2_{\bm{\theta}}\mathcal{L}(\bm{\theta})$.

Instead of optimising the loss function directly, at each iteration of the optimisation process, second-order methods focus on a generating a candidate update $\Delta\bm{\theta}$  through minimising Eqn. \eqref{eq:2ndtaylor} where $\mathbf{H}$ is approximated by a candidate matrix $\mathbf{B}$. Differentiating  Eqn. \eqref{eq:2ndtaylor} and setting it to zero yields the {\it Newton direction} 
\begin{equation}
\Delta\bm{\theta}=-\mathbf{B}^{-1}\nabla_{\bm{\theta}} \mathcal{L}(\bm{\theta}).
\label{eq:newtondir}
\end{equation}
However, computing this direction directly is  expensive since it requires $\mathbb{O}(D^2)$ complexity to store $\mathbf{B}$ and $\mathbb{O}(D^3)$ to invert it.
These obstacles, however, can be overcome by employing inexact Newton methods such as the CG algorithm.

\begin{algorithm}
\caption{The linear conjugate gradient (CG) algorithm.}
\label{CGalgo}
\begin{algorithmic}
\State Let $M$ be the number of CG iterations to execute
\State Set $\bm{v}_{0}\gets-\nabla_{\bm{\theta}} \mathcal{L}(\bm{\theta})$, $\bm{r}_{0}\gets\bm{v}_{0}, m\gets0$
\While {$m<M$}
\State Compute $\bm{r}_{m}^{\text T}\bm{r}_{m}$
\State Set $\alpha_{m}\gets{\bm{r}_{m}^{\text T}\bm{r}_{m}}/{\bm{v}_{m}^{\text T}\mathbf{B}\bm{v}_{m}}$
\State Update $\Delta\bm{\theta}_{m+1}\gets\Delta \bm{\theta}_{m}+\alpha_{m}\bm{v}_{m}$
\State Update $\bm{r}_{m+1}\gets\bm{r}_m-\alpha_{m}\mathbf{B}\bm{v}_{m}$
\State Compute $\bm{r}_{m+1}^{\text T}\bm{r}_{m+1}$
\State Set $\beta_{m+1}\gets{\bm{r}_{m+1}^{\text T}\bm{r}_{m+1}}/{\bm{r}_{m}^{\text T}\bm{r}_{m}}$
\State Update $\bm{v}_{m+1}\gets\bm{r}_{m+1} + \beta_{m+1}\bm{v}_{m}$
\EndWhile
\Return $\Delta{\bm{\theta}}$ as the one that leads to the best performance on the validation set among $\Delta{\bm{\theta}}_1,\Delta{\bm{\theta}}_2,\ldots,\Delta{\bm{\theta}}_M$
\end{algorithmic}
\end{algorithm}

\subsection{The CG Algorithm}
 \label{ssec:CGalg}
 CG is an iterative algorithm that implicitly minimises the quadratic function
\begin{equation}
    g(\Delta\bm{\theta}_m)=\dfrac{1}{2} \Delta\bm{\theta}^\text{T}_m\mathbf{B}\Delta\bm{\theta}_m+\Delta\bm{\theta}^{\text T}_m\nabla_{\bm{\theta}}\mathcal{L}(\bm{\theta})
    \label{eq:CGball}
\end{equation} 
 by solving the linear linear system 
\begin{equation}
 \label{eq:linearsys}
 \mathbf{B}\Delta\bm{\theta}=-\nabla_{\bm{\theta}} \mathcal{L}(\bm{\theta}),
 \end{equation}
 At each iteration $m$, the algorithm minimises Eqn. \eqref{eq:CGball} by taking an appropriate step size $\alpha_m$  along a conjugate search direction  $\bm{v}_m$ w.r.t. $\mathbf{B}$  such that the direction is never revisited at subsequent iterations. When $\mathbf{B}$ is  symmetric and  positive definite, the solution to the linear system yields a unique minimiser of Eqn.~\eqref{eq:CGball}.  Since Eqn.~\eqref{eq:CGball} only approximates $\mathcal{L}(\bm{\theta})$, the standard practice in training parametric models such as NN is to run only finite iterations of the algorithm in which minimising the quadratic function correlates with reductions in the empirical loss \citep{Martens2020}.

The detailed CG procedure is presented as Algorithm \ref{CGalgo}, for which an excellent explanation can be found in \citep{shewchuk1994introduction}.
The key features of CG are summarised below.
\begin{itemize}
\item $\bm{v}_1,\bm{v}_2,\ldots,\bm{v}_M$ are $\mathbf{B}$-orthogonal. This means any direction $\bm{v}_m$ is conjugate to any other direction w.r.t. $\mathbf{B}$. 
\item CG computes $\mathbf{B}\,\bm{v}_{m}$ instead of the Hessian matrix itself. When $\mathbf{B}$ is chosen to approximate the Hessian matrix, the method is known as Hessian free \citep{Martens2010,bottou2016optimization}.
\item Since $\Delta\bm{\theta}_{M}=-\alpha_0\nabla_{\bm{\theta}} \mathcal{L}(\bm{\theta})+\sum\nolimits_{m=1}^{M-1}\alpha_m\bm{v}_m$, $\Delta\bm{\theta}_0$ equals $-\alpha_0\nabla_{\bm{\theta}}\mathcal{L}(\bm{\theta})$ and can be seen as the update obtained by gradient descent with an optimal learning rate that minimises  Eqn.~\eqref{eq:CGball}. 
\item CG will converge monotonically to the exact Newton direction within $M$ iterations, if $\mathbf{B}$ has $M$ distinct or clustered eigenvalues \citep{NocedalBook}. 
\end{itemize}

\subsection{Approximating the Hessian with the  Gauss-Newton Matrix}
\label{ssec:GNmatrix}
This section reviews using the GN matrix as $\mathbf{B}$ in the approximation of $\mathbf{H}$  in the HF method  \citep{martens2011learning}. 
For simplicity, here we consider the case with only one input sample, a frame at time $t$. It is straightforward to generalise the method and equations to the case of many input samples.
Let $\bm{a}^{\text{out}}_t=\{a_{t,1},a_{t,2},\ldots,a_{t,K}\}$ be the logit values (the input values to the softmax output activation function), $K$ be the output layer size, $\mathbf{H}_{ij}$ be the element of the $i$\,th row and $j$\,th column of $\mathbf{H}$, $\mathbf{H}_{ij}$ can be written as
\begin{align}
\label{eq:Hij}
&\mathbf{H}_{ij}=\dfrac{\partial}{\partial\theta_j}\left(\dfrac{\partial\mathcal{L(\bm{\theta})}}{\partial\theta_i}\right)\\
\nonumber&=\sum_{k=1}^{K}\dfrac{\partial a_{t,k'}}{\partial\theta_j}\sum_{k'=1}^{K}\dfrac{\partial a_{t,k}}{\partial\theta_i}\dfrac{\partial^2\mathcal{L}(\bm{\theta})}{\partial a_{t,k}\partial a_{t,k'}}+\sum_{k=1}^{K}\dfrac{\partial\mathcal{L}(\bm{\theta})}{\partial a_{t,k}}\dfrac{\partial^2 a_{t,k}}{\partial\theta_i\partial\theta_j}
\end{align}
In Eqn.~\eqref{eq:Hij}, the first term can be interpreted as the contribution to the Hessian made by the variation in $\bm{a}^{\text{out}}_t$, while the second term is the contribution due to the variation in $\bm{\theta}$. If $\bm{\theta}$ is around a region of local minimum {w.r.t.} the average empirical loss over samples, then $\partial\mathcal{L}(\bm{\theta})/\partial a_{t,k}\approx0$ and the second term is negligible. As a result,
\begin{align}
\label{eq:Gij}
    \mathbf{H}_{ij}\approx\sum_{k=1}^{K}\dfrac{\partial a_{t,k'}}{\partial\theta_j}\sum_{k'=1}^{K}\dfrac{\partial a_{t,k}}{\partial\theta_i}\dfrac{\partial^2\mathcal{L}(\bm{\theta})}{\partial a_{t,k}\partial a_{t,k'}}=\mathbf{G}_{ij},
\end{align}
which is an element of the GN matrix, $\mathbf{G}$.
By rearranging Eqn.~\eqref{eq:Gij}, the GN matrix can be written as
\begin{align}
\mathbf{G}=\mathbf{J}^\text{T}(\nabla^2_{\bm{a}^{\text{out}}_t}\mathcal{L}(\bm{\theta}))\,\mathbf{J},
\label{eq:GNMatrix}
\end{align}
where $\mathbf{J}$ is the Jacobian matrix  $\nabla_{\bm{\theta}}(\bm{a}^{\text{out}}_t)$,
and $\nabla^2_{\bm{a}^{\text{out}}_t}\mathcal{L}(\bm{\theta})$ is the Hessian matrix w.r.t. $\bm{a}^{\text{out}}_t$. 

As shown by \citet{schraudolph2002fast}, for matching  loss  functions  where  $\nabla^2_{\bm{a}^{\text{out}}_t}\mathcal{L}(\bm{\theta})$ is positive definite {w.r.t.} the NN output logits, the GN matrix is guaranteed to be positive semi-definite. When the GN approximation is applied to  to lattice-based MBR training \citep{kingsbury2012}, the component $\nabla^2_{\bm{a}^{\text{out}}_t}\mathcal{L}_{\text{MBR}}(\bm{\theta})$ which we denote as $\hat{\mathbf{H}}$ takes the following form \citep{AdnanHaider}:
\begin{align}
\label{eq:MBRHessian}
\hat{\mathbf{H}}={\kappa^2}\left(\text{diag}(\bm {\gamma}^{\text{MBR}}_t)-\bm{\gamma}^{\text{MBR}}_t\bm{\gamma}^{\text T}_t\right),
\end{align}
where $\kappa$ is the acoustic scaling factor, $\gamma_{t,k}$ and $\gamma^{\text{MBR}}_{t,k}$ are correspondingly the ML and MBR occupancy at time $t$ {w.r.t.} the HMM state $k$ (tied to the DNN output unit $k$), and $\text{diag}(\cdot)$ converts a vector into a diagonal matrix.

Regarding ML training in the ASR literature, the term ``occupancy'' often refers to $\gamma_{t,k}=P_{\bm{\theta}}(q_t=k|\mathbf{O},\mathbf{W}^{\text{ref}})$, the posterior probability showing how probable frame $t$ is aligned with state $k$ given a pair of sequences $\mathbf{O}$ and $\mathbf{W}^{\text{ref}}$. $\gamma_{t,k}$ is often calculated using the \textit{forward-backward} procedure \citep{Baum1967}. 
Regarding MBR sequence training, the occupancy $\gamma^{\text{MBR}}_{t,k}$ is defined based on the gradients of the loss function ${\partial\mathcal{L}_{\text{MBR}}(\bm{\theta})}/{\partial a_{t,k}}=-\kappa\gamma^{\text{MBR}}_{t,k}$, where 
$\gamma^{\text{MBR}}_{t,k}=\gamma_q(c_q-c_{\text{avg}})\,\gamma_{t,k}$, $\gamma_q$ is the occupancy passing through arc $q$, $c_{\text{avg}}$ and $c_q$ are the weighted average correctness of all hypotheses and the hypotheses including arc $q$. $\gamma_q$, $c_q$, and $c_{\text{avg}}$ can be collected by performing a modified forward-backward procedure to align every arc in the lattice with $\mathbf{O}$ \citep{Povey2005}. 

From Eqn.~\eqref{eq:MBRHessian}, it can seen that $\nabla^2_{\bm{a}^{\text{out}}_t}\mathcal{L}_{\text{MBR}}(\bm{\theta})$ is not positive definite {w.r.t.} the NN output logits and hence the GN matrix is no longer guaranteed to be positive semi-definite. Interestingly, even through the GN matrix  no longer possesses the property of being positive semi-definite, its use as an approximation to the Hessian has been shown empirically to be 
effective in obtaining  stable WER reductions from lattice-based MBR sequence training \citep{kingsbury2012,Sainath2013b,Sainath2013a,Dognin2013}. The next section describes a particular property of the GN matrix that has been recently shown by \citet{haider2018} to be effective when using  a disciminative sequence criterion to train NN models with sharp softmax distributions.

\subsection{Scaling by the Gauss-Newton Matrix}
\label{ssec:GNscaling}

The frame-level CE loss used to train an NN acoustic model from random initialisation using frame-to-HMM-state alignments often results in distribution of the NN softmax outputs being very sharp, in particular for models that use ReLU activation functions. 
 This was reported in \citep{AdnanHaider} to be a contributing factor in achieving only very small improvements from discriminative sequence training.
A similar issue has been observed in connectionist temporal classification \citep{CTC}, where the sharp distributions are caused by the blank unit 
instead of the 0-1 training labels. 
According to Eqn.~\eqref{eq:Gij}, $\mathbf{G}$ captures the curvature of the training loss w.r.t. the model output distribution. Re-scaling the gradients by $\mathbf{G}^{-1}$ effectively de-weights the back-propagated information that can induce large changes in the loss value in EBP. 
In the context of discriminative sequence training, this regularises the sharp model output distributions and improves the performance of MBR training \citep{AdnanHaider}.

\subsection{ Matrix-Vector-Products with the Gauss-Newton Matrix}
\label{ssec:mbr}

Within each iteration of CG, a multiplication of the GN matrix with a vector $\bm{v}$ ($\mathbf{B}\,\bm{v}_{m}$ in Alg.~\ref{CGalgo}) in the parameter space, $\mathbf{G}\,\bm{v}=\mathbf{J}^{\text T}(\nabla^2_{\bm{a}^{\text{out}}_t}\mathcal{L}(\bm{\theta}))\,\mathbf{J}\,\bm{v}$, corresponds to the following sequential multiplication:
\begin{itemize}
\item Computing the directional derivative $\mathbf{J}\,\bm{v}$ using a modified forward propagation procedure.
\item Multiplying the resulting vector $\mathbf{J}\,\bm{v}$ by
 $\nabla^2_{\bm{a}^{\text{out}}_t}\mathcal{L}(\bm{\theta})$ which  corresponds to $\hat{\mathbf{H}}$ in this context.
\item Since the error backpropgation (EBP) procedure computes $\mathbf{J}^{\text T}(\nabla_{\bm{a}^{\text{out}}_t}\mathcal{L}(\bm{\theta}))$, $\mathbf{G}\,\bm{v}$ can be obtained using EBP by
replacing $\nabla_{\bm{a}^{\text{out}}_t}\mathcal{L}(\bm{\theta})$ with $(\nabla^2_{\bm{a}^{\text{out}}_t}\mathcal{L}(\bm{\theta}))\,\mathbf{J}\,\bm{v}$.
\end{itemize}

To compute $\mathbf{J}\,\bm{v}$ efficiently using a modified forward propagation procedure, 
\citet{pearlmutter1994fast} introduced an operator $\mathcal{R}(\cdot)$ to calculate the directional derivative $\nabla_{\bm{\theta}}(\cdot)\,\bm{v}$, and there is $\mathcal{R}(\bm{\theta})=\bm{v}$.
For a fully-connected (FC) layer with $a_{t,k}=\sum_{j}u_{kj}x_{t,j}+b_k$, where $u_{kj}$ is the weight value associated with the $j$\,th input unit and $k$\,the output unit of the layer, $x_{t,j}$ and $b_j$ are the $j$\,th elements of the input and bias vectors, it is easy to show that
\begin{align}
    \nonumber&\mathcal{R}(a_{t,k})=\sum\nolimits_j\mathcal{R}(u_{kj})\,x_{t,j}+\sum\nolimits_ju_{kj}\mathcal{R}(x_{t,j})+\mathcal{R}(b_j)\\
    \label{eq:Rfwd}
    &=\sum\nolimits_jv_{kj}\,x_{t,j}+\sum\nolimits_ju_{kj}h'(a_{t,j})\mathcal{R}(a_{t,j})+v_j,
\end{align}
where $v_{kj}$ and $v_j$ are the elements corresponding to $u_{kj}$ and $b_j$ of $\bm{v}$, 
$h(\cdot)$ is the hidden activation function transforming $a_{t,j}$ to $x_{t,j}$ by $x_{t,j}=h(a_{t,j})$, $a_{t,j}$ is an activation value produced by a previous layer. According to Eqn.~\eqref{eq:Rfwd}, $\mathcal{R}(\bm{a}^{\text{out}}_t)$ can be calculated efficiently by modifying the forward propagation procedure, which results in the required directional derivative $\mathbf{J}\,\bm{v}$ since $\mathcal{R}(\bm{a}^{\text{out}}_t)=\nabla_{\bm{\theta}}(\bm{a}^{\text{out}}_t)\,\bm{v}=\mathbf{J}\,\bm{v}$. A detailed explanations of the modified forward procedure can be found in \citep{BishopPRMF}.

In addition, it is not necessary to compute and store $\hat{\mathbf{H}}$ explicitly. For lattice-based MBR training,  $\hat{\mathbf{H}}\,\mathcal{R}(\bm{a}^{\text{out}}_t)$ can be directly calculated by
\begin{align*}
\hat{\mathbf{H}}\,\mathcal{R}(\bm{a}^{\text{out}}_t)
=\kappa^2{\gamma}_t \odot \mathcal{R}(\bm{a}^{\text{out}}_t)-\kappa^2\bm{\gamma}^{\text{MBR}}_t \left ( \bm{\gamma}^{\text T}_t \mathcal{R}(\bm{a}^{\text{out}}_t)\right),
\end{align*}
where $\odot$ refers to the Hadamard product.

\section{Improving CG for Distributed Training}
\label{sec:ImprovingHF}
The CG-based distributed optimisation framework is presented in this section, which can be used for HF \citep{Martens2010,kingsbury2012}. 
In practice, HF is found to have a high computational cost since a large number of CG iterations are required to perform in sequence to find an effective update \citep{Sainath2013b}.
In this section, a modification to standard CG is presented that overcomes the numerical instability issue, which can reduce the number of CG iterations required by a factor of about twenty. 
Furthermore, a gradient normalisation method is proposed to improve the performance of CG for shared parameters, which facilitates the training of convolutional and recurrent models.

\subsection{CG-based Distributed Optimisation}
\label{ssec:gbdist}
\begin{figure}[h]
    \centering
    \includegraphics[width=1.0\linewidth]{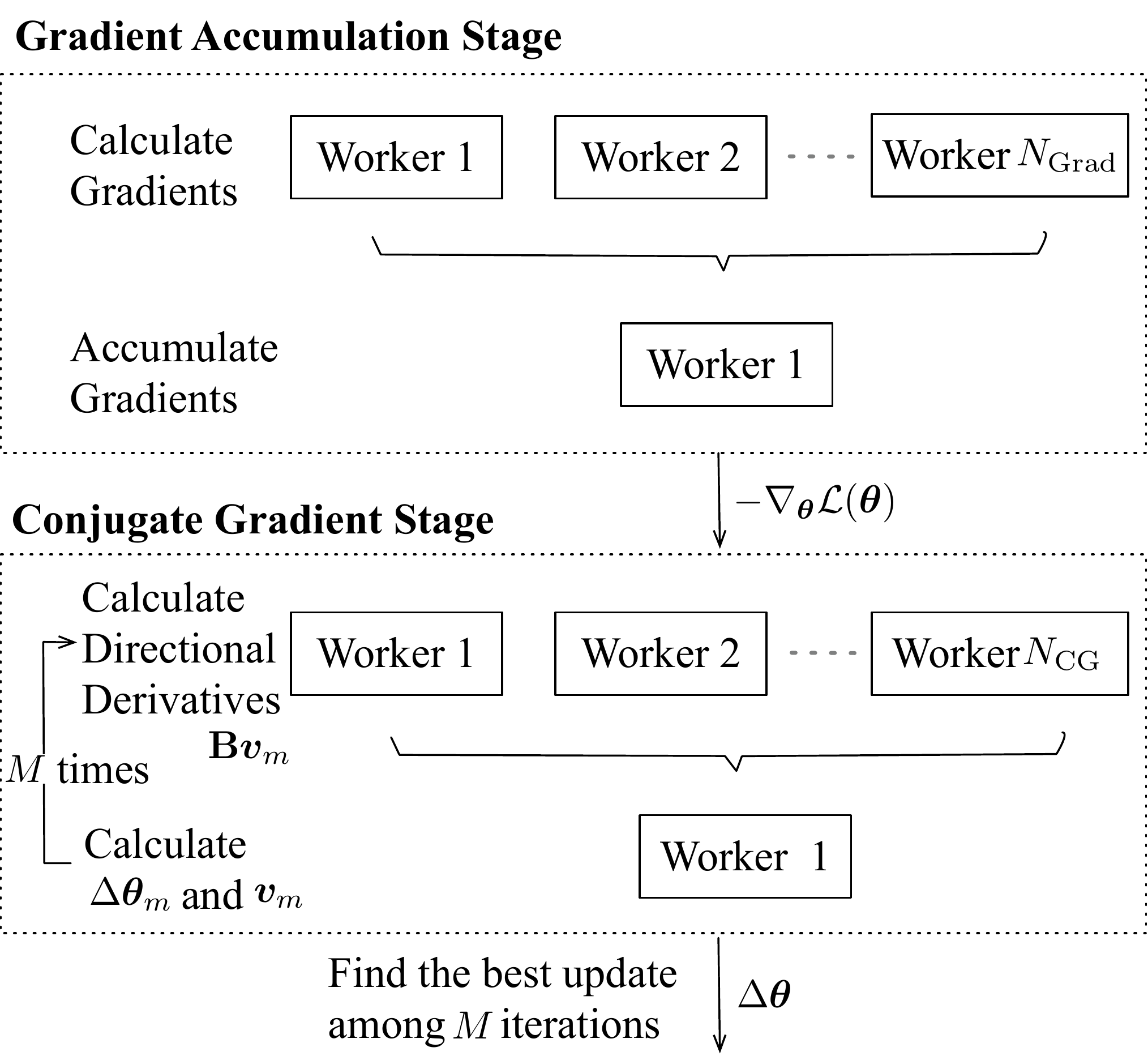}
    \caption{A flow chart of our CG-based distributed optimisation framework, where $N_{\text{Grad}}$ and $N_{\text{CG}}$ are the utterance numbers in the gradient batch and CG batch respectively. They are also the maximum number of workers allowed for gradient calculation and directional derivative calculation. $\bm{v}_m$ and $\Delta\bm{\theta}_m$ are the conjugate direction and the proposed parameter update at the $m$\,th CG iteration.}
    \label{fig:framework}
    \vspace{-0.5cm}
\end{figure}

As shown in Fig.~\ref{fig:framework}, the CG-based optimisation framework consists of two stages: the gradient accumulation stage and the CG stage.
The gradient accumulation stage approximates the true gradient $\nabla_{\bm{\theta}}\mathcal{L}(\bm{\theta})$ with an average of the gradients computed over every sample in a large data batch,
which is referred to as a \textbf{gradient batch}. Here, a sample is an utterance for discriminative sequence training. 
In this stage, most of the calculations are used to compute the gradient w.r.t. each sample using the forward propagation and EBP procedures. These can be conducted in parallel using multiple workers. The negative gradients are then accumulated and averaged to form $\bm{v}_0$ for the CG stage. 

In the CG stage, a sequence of CG iterations is used to find the parameter update $\Delta\bm{\theta}$ with another batch of samples called the \textbf{CG batch}. As explained in Section~\ref{ssec:GNmatrix}, $\mathbf{G}\,\bm{v}_m$ is calculated at each CG iteration $m$ using the modified forward propagation and EBP. Such calculations can be conducted in parallel using a separate worker for each sample in the CG batch, whose output statistics can be accumulated for the rest of the steps of Alg.~\ref{CGalgo}.
The updated search direction $\bm{v}_{m+1}$ will monotonically improve upon $\bm{v}_m$ in terms of the loss value of the CG batch.  After a certain number of iterations, $\Delta\bm{\theta}_m$ with the best training loss performance on the CG mini-batch is returned as the direction found by the CG.  

In this paper, executing the two stages once is referred to as an \textbf{update} since it is used to find one $\Delta\bm{\theta}$ to update the current parameter $\bm{\theta}$. In practice, we divide the whole training set randomly into $C$ partitions, with each of them being used as a gradient batch for an update, and hence each training epoch comprises $C$ updates performed in sequence. The CG batch is often much smaller than the gradient batch since it often needs to be processed for many iterations in the CG stage. 
In our experiments, we found it is better to sample the CG batch from the entire training set rather than just from the corresponding gradient batch. 

\subsection{Improving the Stability of CG}
\label{ssec:cgstab}
Although the calculation of an individual CG iteration can be distributed over many workers, CG iterations are still required to be performed  sequentially. 
As reported in previous studies \citep{Martens2010,wiesler2013,Martens2015}, the HF method often requires about 200 CG iterations to find an effective update for training a DNN, even in the case of lattice-based discriminative sequence training \citep{kingsbury2012,Sainath2013b}.
This means that, in practice, CG restricts the training speed. Next, we explain the cause of this issue, and a solution is proposed to improve the stability of CG, which will be shown in Sec. \ref{sec:ExpResults} to yield effective updates only from few iterations of CG.

As discussed in Sec.~\ref{ssec:GNmatrix}, for a matching loss function, such as the CE loss together with the softmax function, 
the GN matrix $\mathbf{G}$ is in theory guaranteed to be positive semi-definite. However, even in our CE training experiments, it was observed that $\mathbf{G}$ could at times be negative \citep{AdnanHaider}. 
This issue was found to be a result of insufficient arithmetic accuracy when calculating the directional derivatives $\mathbf{J}\,\bm{v}_m$ for a CG iteration $m$. More specifically, let $\|\cdot\|_2$ be the L2 norm of a vector, when $\|\bm{\theta}\|_2\gg\|\bm{v}_m\|_2$, Eqn.~\eqref{eq:Rfwd} becomes
\begin{align*}
\mathcal{R}(a_{t,k})\approx\sum\nolimits_ju_{kj}h'(a_{t,j})\mathcal{R}(a_{t,j}),
\end{align*}
due to the limited precision of the floating-point arithmetic, which can lead to an incorrect value of $\mathbf{G}$ that may no longer be a positive semi-definite matrix. 

For large scale distributed training \citep{Sainath2013b}, the issue of negative $\mathbf{G}$ is resolved by using Tikhonov damping \citep{tikhonov1998nonlinear}, which uses $\mathbf{G}+\eta\,\mathbf{I}$ instead of $\mathbf{G}$ in CG. This corresponds to taking comparatively more conservative steps along the individual conjugate directions, and considerably slows down training  as more CG updates are required to get a good overall solution.
In the scenario when $\eta$ is large, Tikhonov damping is effectively  analogous to an SGD step.
Instead of Tikhonov damping, we propose to modify each CG iteration by using 
\begin{align*}
    \bm{v}'_m=({\|\bm{\theta}\|_2}/{\|\bm{v}_m\|_2})\,\bm{v}_m
\end{align*}
to compute $\mathbf{J}\,\bm{v}'_m$. Afterwards, $\mathbf{J}\,\bm{v}_m$ is obtained by
\begin{align*}
    \mathbf{J}\,\bm{v}_m=({\|\bm{v}_m\|_2}/{\|\bm{\theta}\|_2})\,\mathbf{J}\,\bm{v}'_m.
\end{align*}
In our experiments, it was found that this improved CG algorithm can often produce an effective $\Delta\bm{\theta}$ with about 8 iterations. 

\subsection{Improving CG for Shared Parameters \label{sec:CGpredcond}}
In this section, it is demonstrated how CG can be adapted to perform efficiently for models with shared parameters, such as the TDNN \citep{timedelayWaibel1,DanTDNN,tdnns2018icassp} and LSTM \citep{Hochreiter1997,Graves2013,Sak2014},
both widely used for acoustic modelling. In contrast to a DNN, a TDNN uses a sequence of fully FC layers to perform 1-dimensional (-dim) convolutions across time, whose input vectors the concatenation of $\bm{x}_{t_1}$ and $\bm{x}_{t_2}$, the output from their direct preceding layers of two different time steps $t_1$ and $t_2$. That is,
\begin{align*}
    \bm{y}_t=h(\mathbf{U}\,\text{Concat}(\bm{x}_{t_1},\bm{x}_{t_2})+\bm{b}),
\end{align*}
where $\text{Concat}(\cdot)$ is the concatenation operation; $h(\cdot)$, $\mathbf{U}$ and $\bm{b}$ are the activation function, weight matrix and bias vector of the FC layer. Hence, the directional derivatives of a TDNN are also calculated using Eqn.~\eqref{eq:Rfwd}.
Alternatively, a TDNN can be viewed as a feedforward model with a binary tree structure by duplicating each FC layer for the relevant time steps, and the parameters $\mathbf{U}$ and $\bm{b}$ are shared across time. 

Regarding an (Elman network) RNN \citep{rumelhart,elmanRNN}, $\bm{h}_t$, the output vector at time $t$, is generated by transforming a concatenation of $\bm{h}_{t-1}$ and the current input $\bm{x}_t$ with an FC layer whose parameters are $\mathbf{U}$ and $\bm{b}$, \textit{i.e.} $\bm{y}_t=f(\mathbf{U}\,\text{Concat}(\bm{x}_t,\bm{h}_{t-1})+\bm{b})$. 
An LSTM is an improved RNN with enhanced long-term memory capability based on the gating mechanism, which uses $\bm{i}_t$, $\bm{f}_t$, and $\bm{o}_t$, the sigmoidal output vectors from three extra FC layers, to simulate the logic gates of a memory circuit to maintain a memory cell $\bm{c}_t$. Given the parameters of the three extra FC layers, $\mathbf{U}_i$ and $\bm{b}_i$, $\mathbf{U}_f$ and $\bm{b}_f$, and $\mathbf{U}_o$ and $\bm{b}_o$, an LSTM layer can be specifically presented as    
\begin{align*}
	\bm{i}_t&= {\sigma}(\mathbf{U}_i\,\text{Concat}(\bm{x}_t,\bm{h}_{t-1})+\bm{b}_i),\\
	\bm{f}_t&= {\sigma}(\mathbf{U}_f\,\text{Concat}(\bm{x}_t,\bm{h}_{t-1})+\bm{b}_f),\\
	\bm{c}_t&=\bm{f}_t\odot\bm{c}_{t-1}+\bm{i}_t\odot{\tanh}(\mathbf{U}\,\text{Concat}(\bm{x}_t,\bm{h}_{t-1})+\bm{b}),\\
	\bm{o}_t&= {\sigma}(\mathbf{U}_o\,\text{Concat}(\bm{x}_t,\bm{h}_{t-1})+\bm{b}_o),\\
	\bm{h}_t&=\bm{o}_t\odot{\tanh}(\bm{c}_t),
\end{align*}
where $\sigma(\cdot)$ and $\tanh(\cdot)$ are the sigmoid and hyperbolic tangent activation functions. An LSTM layer can be implemented with four FC layers and the Hadamard product. Therefore, the directional derivative for an LSTM can be calculated using Eqn.~\eqref{eq:Rfwd} and the following rule of the $\mathcal{R}(\cdot)$ operator for gating:
\begin{align}
  \label{eq:R4gating}
  \mathcal{R}\big(\bm{g}_{t}\odot\bm{z}_{t}\big)&=\mathcal{R}\big(\bm{g}_{t}\big)\odot\bm{z}_{t}+\bm{g}_{t}\odot\mathcal{R}\big(\bm{z}_{t}\big),
\end{align}
where $\bm{g}_t$ and $\bm{z}_t$ are two example vectors.
Further by unfolding through time for $u$ steps \citep{robinson1987,Werbos1988}, a folded LSTM layer becomes $u$   unfolded layers with feedforward connections sharing all of their parameters.
The input of the $v$\,th feedforward layer is a concatenation of the output from the $(v-1)$\,th layer and $\bm{x}_{t-u+v}$.


Next we present a modification to the CG algorithm to make the algorithm more effective for models with shared parameters. From Alg.~\ref{CGalgo}, both the step size $\alpha_m$ and the conjugate search direction $\bm{v}_m$ are determined by the dot product of the residual $\bm{r}^{\text T}_{m}\bm{r}_{m}$ and the directional derivative $\mathbf{G}\,\bm{v}_m$. For models such as the TDNN and RNN, shared parameters receive more updates and hence will contribute more to the norm of the vectors $\bm{r}_0$ and $\mathbf{G}\,\bm{v}_m$ than the parameters which are not shared. Careful preliminary experiments using TDNNs and LSTMs found that  in situations where the shared parameters dominate the norm of these vectors, the CG algorithm  was slow to find an update direction that could reduce the loss value. Our solution is apply a diagonal scaling to  $\bm{r}_m$ and $\mathbf{G}\,\bm{v}_m$ by a matrix whose diagonal entries correspond to the square root of the number of times that a parameter is shared. This ensures that the L2 norm of the vectors are not dominated by the contributions of the shared parameters and enables a more effective update for the other parameters.
Such an  approach corresponds  to preconditioning the CG algorithm \citep{shewchuk1994introduction} 
by applying the diagonal scaling only to $\bm{r}_0$ among all the residuals $\bm{r}_m$. 


By using the algorithm given in \citep{zhang:2015HTK}, the diagonal scaling can be efficiently achieved by adding an operation to normalise the resulting gradients or directional derivatives at the end of the EBP procedure by  the number  times a parameter is shared. Experiments showed that this solution enabled CG to find progressively better update directions in each CG iteration for TDNNs and LSTMs \citep{AdnanHaider}. More details of the theoretical analysis and experimental evidence can be found in \citep{AdnanHaider}.

\section{Natural Gradient Optimisation}
\label{sec:NaturalGrad}
In this section first NG is reviewed. Then the proposed CG-based optimisation framework for NG for discriminative sequence training is presented.
\subsection{Natural Gradient Descent}
\label{ssec:NaturalGrad}
This section presents an overview of NG descent \citep{Amari1997,pascanu2013}. 
Let  $\mathcal{M}$ denote the space of all  temperature-modulated probability distributions $P_{\bm{\theta}}(\mathbf{W}|\mathbf{O})$\footnote{ $P_{\bm{\theta}}(\mathbf{W}|\mathbf{O})$ is determined by a chosen value of  $\kappa$ (see Eqn.~\eqref{eq:MMIloss}).}. 
In the context of optimisation,
NG tries to find an update $\Delta\bm{\theta}$ that minimises the loss function locally while keeping a similar probability distribution to the one resulting from
the current $\bm{\theta}$. 
That is,
\begin{align}
    \label{eq:ngdef}
	&\Delta\hat{\bm{\theta}}= \arg\min_{\Delta\bm{\theta}}  \mathcal{L}(\bm{\theta}+\Delta\bm{\theta}), \\
	\nonumber&\mbox{s.t. }  \mathbb{E}_{p(\mathbf{O})}\left[{\text{KL}}\left ({P}_{\bm{\theta}}(\mathbf{W}|\mathbf{O})\| {P}_{\bm{\theta}+\Delta\bm{\theta}}(\mathbf{W} |\mathbf{O})\right)\right]\leqslant\varepsilon
\end{align}
where KL$(\cdot||\cdot)$ refers to the KL-divergence 
 and $\varepsilon$ is a constant controlling the speed of exploration along the manifold.  In Eqn.~\eqref {eq:ngdef}, we assume  $p_{\bm{\theta}}(\mathbf{W},\mathbf{O})=P_{\bm{\theta}}(\mathbf{W}|\mathbf{O})p(\mathbf{O})$.

Eqn.~\eqref{eq:ngdef} can be formulated as an equivalent constrained optimisation problem in the parameter space by using 
\textit{Lagrange multipliers}, which restrict the exploration of the parameter space to be within a local neighbourhood of the current estimate $\bm{\theta}$. Approximating $\mathcal{L}(\bm{\theta}+\Delta\bm{\theta})$ by its first-order Taylor approximation $\mathcal{L}(\bm{\theta})+\Delta\bm{\theta}^{\text T}\nabla_{\bm{\theta}}\mathcal{L}(\bm{\theta})$, and the KL-divergence constraint by its second-order Taylor approximation \citep{AmariBook} yields
\begin{equation}
\setlength{\mathindent}{0cm}
 \label{eq:NGlagrangian}
 \Delta\hat{\bm{\theta}}=\arg\min_{\Delta\bm{\theta}}\left\{ \mathcal{L}(\bm{\theta})+\Delta \bm{\theta}^{\text T}\nabla_{\bm{\theta}}\mathcal{L}(\bm{\theta})+\frac{\lambda}{2} \Delta \bm{\theta}^{\text T} \mathbf{F}\Delta\bm{\theta}\right\},
\end{equation}
where $\lambda$ is the Lagrange multiplier that controls the compromise between 
minimising the loss and satisfying the KL-divergence constraint. $\mathbf{F}$ is referred to as the FI matrix.
\begin{equation}
\setlength{\mathindent}{0cm}
\label{eq:FI}
\mathbf{F}=\mathbb{E}_{p_{\bm{\theta}}(\mathbf{W}|\mathbf{O})}\left[\nabla_{\bm{\theta}}\ln P_{\bm{\theta}}(\mathbf{W}|\mathbf{O})\nabla_{\bm{\theta}}\ln P_{\bm{\theta}}(\mathbf{W}|\mathbf{O})^{\text T}\right],
\end{equation}

Similar to second-order methods, NG 
attempts to minimise a quadratic function at each iteration. 
Differentiating Eqn.~\eqref{eq:NGlagrangian} and equating it to zero yields the solution
\begin{align}
\Delta\bm{\theta}=-\dfrac{1}{\lambda}\,\mathbf{F}^{-1}\nabla_{\bm{\theta}}\mathcal{L}(\bm{\theta}).
\label{eq:NGdir}
\end{align}
This suggests that the update direction obtained by NG transforms the steepest decent direction by taking into account the curvature information of the log-likelihood  given by $\mathbf{F}^{-1}$. It can be shown that such a direction is indeed the optimal descent in the loss surface generated on the manifold $\mathcal{M}$.

Computing the exact FI matrix defined in Eqn.~\eqref{eq:FI} requires taking the expectation over the distribution $p_{\bm{\theta}}(\mathbf{W},\mathbf{O})$ which is infeasible for LVCSR. A standard approach is to approximate this expectation by using the average of samples from $p_{\bm{\theta}}(\mathbf{W},\mathbf{O})$. In this paper, we take samples from the data distribution $p(\mathbf{W},\mathbf{O})$, instead of $p_{\bm{\theta}}(\mathbf{W},\mathbf{O})$. This form of the FI matrix yields the empirical Fisher matrix where the contribution of each utterance corresponds to 
\begin{align}
\label{eq:empiricalFI}
\mathbf{F}&\approx\nabla_{\bm{\theta}}\ln P_{\bm{\theta}}(\mathbf{W}^{\text{ref}} |\mathbf{O})
\nabla_{\bm{\theta}}\ln P_{\bm{\theta}}(\mathbf{W}^{\text{ref}} |\mathbf{O})^{\text T}.
\end{align}
As reviewed in Section~\ref{ssec:reviewnghf}, to reduce the difficulty in calculating $\mathbf{F}^{-1}$, 
$\mathbf{F}$ is often assumed to have a (block) diagonal structure \citep{povey2014parallel,kingma2014adam,Martens2015} with the diagonal blocks corresponding to Kronecker products of two smaller matrices. Meanwhile, CG was proposed to compute the NG direction without calculating $\mathbf{F}^{-1}$ explicitly for both reinforcement learning \citep{Schulman2015} and discriminative sequence training \citep{haider2017}. This will be presented in detail later in Sec.~\ref{ssec:CG4NG}. 

There exist theoretical advantages in using NG. For ML training, assuming  the distribution of the gradient of the expected loss to be an isotopic Gaussian, the NG direction is relevant to a direction in the parameter space  that maximises the probability of reducing the generalisation error~\citep{Roux}. Recently, it has been shown that for convex problems, 
solving Eqn.~\eqref{eq:NGlagrangian} at each iteration results in an exponentially faster convergence speed compared to gradient descent~\citep{bernacchia2018exact}.

\subsection{Natural Gradient with CG for Discriminative Sequence Training}
\label{ssec:CG4NG}
From Eqn.~\eqref{eq:empiricalFI}, the empirical FI matrix $\mathbf{F}$ is guaranteed to be positive semi-definite and therefore the optimisation framework proposed in Sec.~\ref{sec:ImprovingHF} can be used for NG. CG is used to solve $\lambda\,\mathbf{F}\Delta\bm{\theta}=-\nabla_{\bm{\theta}}\mathcal{L}(\bm{\theta})$ (Eqn.~\eqref{eq:NGdir}). 
From Eqn.~\eqref{eq:MMIloss}, $\ln P_{\bm{\theta}}(\mathbf{W}^{\text{ref}}|\mathbf{O})=-\mathcal{L}_{\text{MMI}}(\bm{\theta})$ when $\kappa=1$, and hence Eqn.~\eqref{eq:empiricalFI} can be re-written as
\begin{align}
\label{eq:empiricalFI2}
\nonumber\mathbf{F}&\approx\nabla_{\bm{\theta}}\mathcal{L}_{\text{MMI}}(\bm{\theta})
\nabla_{\bm{\theta}}\mathcal{L}_{\text{MMI}}(\bm{\theta})^{\text T}\\
&=\sum\nolimits_{t=1}^{T}\mathbf{J}^{\text T}\nabla_{\bm{a}^{\text{out}}_t}\mathcal{L}_{\text{MMI}}(\bm{\theta})\nabla_{\bm{a}^{\text{out}}_t}\mathcal{L}_{\text{MMI}}(\bm{\theta})^{\text T}\mathbf{J}.
\end{align}
This uses the fact that $\nabla_{\bm{\theta}}\mathcal{L}_{\text{MMI}}(\bm{\theta})=[\mathbf{J}^{\text T}\nabla_{\bm{a}^{\text{out}}_t}\mathcal{L}_{\text{MMI}}(\bm{\theta})]^{T}_{t=1}$, 
where $\mathbf{J}$ is the Jacobian matrix of $\bm{a}^{\text{out}}_t$ w.r.t. $\bm{\theta}$, and $T$ is the number of frames in the utterance. Thereafter, we denote $\hat{\mathbf{F}}$ as $\nabla_{\bm{a}^{\text{out}}_t}\mathcal{L}_{\text{MMI}}(\bm{\theta})\nabla_{\bm{a}^{\text{out}}_t}\mathcal{L}_{\text{MMI}}(\bm{\theta})^{\text T}$ for simplicity.
Since Eqn.~\eqref{eq:empiricalFI2} has the same form as the GN matrix given in Eqn.~\eqref{eq:GNMatrix}, the procedure given in Sec.~\ref{ssec:mbr} can be used to calculate $\mathbf{F}\,\bm{v}$ (within a CG iteration), which first calculates $\mathbf{J}\,\bm{v}$ using the modified forward propagation, then multiplies the resulting vector by $\hat{\mathbf{F}}$, and at last calculates $\mathbf{J}^{\text T}(\hat{\mathbf{F}}\mathbf{J}\,\bm{v})$ using the EBP procedure. 

Next, it is shown how 
$\hat{\mathbf{F}}$ can be calculated. Similar to ML and MBR training discussed in Sec.~\ref{ssec:mbr}, it is easy to show  $\partial\mathcal{\mathcal{L}_{\text{MMI}}(\bm{\theta})}/\partial a_{t,i}=-\kappa\gamma^{\text{MMI}}_{t,k}$ \citep{Chao2017}, where  $a_{t,k}$ is the logit value at time $t$ of output unit $k$, and $\gamma^{\text{MMI}}_{t,k}$ is the MMI occupancy with $\gamma^{\text{MMI}}_{t,k}=\gamma^{\text{num}}_{t,k}-\gamma^{\text{den}}_{t,k}$. $\gamma^{\text{num}}_{t,k}$ and $\gamma^{\text{den}}_{t,k}$ are the occupancy derived  
separately from the numerator and denominator parts of Eqn.~\eqref{eq:MMIloss}. 

In practice, $\hat{\mathbf{F}}=\kappa^2\gamma^{\text{MMI}}_{t}(\gamma^{\text{MMI}}_{t})^{\text T}$ does not need to be calculated and stored explicitly. 
Recall the $\mathcal{R}(\cdot)$ operator discussed in Sec. \ref{ssec:GNmatrix}, and that $\mathcal{R}(\bm{a}^{\text{out}}_t)=\mathbf{J}\,\bm{v}$, which has the same dimension as $\gamma^{\text{MMI}}_{t}$.
To calculate $\hat{\mathbf{F}}\,\mathbf{J}\,\bm{v}$ directly, $(\gamma^{\text{MMI}}_{t})^{\text T}  \mathcal{R}(\bm{a}^{\text{out}}_t)$ can be obtained first, which is followed by scaling $\gamma^{\text{MMI}}_{t}$. Specifically,
\begin{align*}
 \hat{\mathbf{F}}\,\mathbf{J}\,\bm{v} 
 &=\kappa^2\gamma^{\text{MMI}}_{t} \left((\gamma^{\text{MMI}}_{t})^{\text T}  \mathcal{R}(\bm{a}^{\text{out}}_t)\right).
\end{align*}
The term $(\gamma^{\text{MMI}}_{t})^{\text T}  \mathcal{R}(\bm{a}^{\text{out}}_t)$ is a scalar quantity and can be interpreted as a learning rate for $\gamma^{\text{MMI}}_{t}$.

By comparing Eqns.~\eqref{eq:2ndtaylor} and \eqref{eq:NGlagrangian}, the  difference between the  NG and HF approach  when applied to minimise any arbitrary smooth loss function  lies in the matrix used in the second-order term, $\mathbf{G}$ and $\mathbf{F}$. For HF, the GN matrix $\mathbf{G}$ requires calculating the Hessian of the MBR loss w.r.t. the logit values. For NG, the empirical FI matrix is calculated as the outer product of the gradients of the MMI loss w.r.t. the logit values, regardless of the loss used for training. In the case of MBR training, NG provides an efficient procedure to combine MBR loss with MMI loss, which is similar to the widely used ``MMI prior'' method for GMM-HMM MBR training that interpolates MBR with MMI in the loss function \citep{htk,Zhang2017}.

\section{Regulating NG Updates}
\label{sec:NGHF}

In Sec.~\ref{sec:NaturalGrad} it is shown how NG descent corresponds to scaling and rotating the gradient $\nabla_{\bm{\theta}}\mathcal{L}(\bm{\theta})$ through multiplication with the inverse $\mathbf{F}$ matrix.  In a scenario where the training criterion  is only an approximation of the evaluation metric, it is shown in \citep{haider2018,AdnanHaider}  that both the approximate NG descent and SGD can at times follow a path in the parameter space where  generalisation improvements {w.r.t.} the training criterion on the validation set  fail to correlate with reductions in WER (the evaluation criterion of interest). Thus in such training paradigms, over-fitting can occur not only due  to the lack of training data but also  due to the underlying criterion mismatch. 
Depending on the task, it will be attractive to have a mechanism that regulates the amount of scaling and rotation of the loss gradient to achieve good generalisation. The following subsection presents a common framework to regulate the NG direction or the gradient descent direction by using an appropriate  choice of $\mathbf{B}$. The procedure presented here relies on a re-derivation of Taylor's theorem using  the concepts of  manifolds, tangent vectors and directional derivatives  from the perspective of differential geometry. The derivation is provided in detail in the technical report \citep{haider2018common}. An overview of the necessary underlying concepts can be found in \citep{AmariBook,deFelice:1990hu}.

\subsection{A Common Framework to Regulate Natural Gradient and Gradient Descent}
 Assuming that  the loss function $\mathcal{L}(\bm{\theta})$ is sufficiently smooth, using Taylor's theorem, second order methods  proceed to minimise the loss by  minimising the following function at each iteration $m$:
\begin{align*}
\setlength{\mathindent}{0cm}
\Delta\bm{\theta}' =\argmin_{\Delta\bm{\theta}}\left\{
\mathcal{L}(\bm{\theta}_m)+\Delta\bm{\theta}^{\text T}\nabla_{\bm{\theta}}\mathcal{L}(\bm{\theta}_m) + \dfrac{1}{2} \Delta\bm{\theta}^\text{T}\mathbf{B}\Delta\bm{\theta}\right\}
\end{align*}
where $\bm{\theta}_m$ corresponds to the current parameter estimate. Using the fundamental theorem of calculus, in \citep{haider2018common}, it is shown that solving the above problem can be cast as an equivalent  minimisation problem in  the tangent space of the current parameter estimate  $\mathcal{T}({\bm{\theta}_m})$:
\begin{align*}
\argmin_{\Delta\bm{\theta}\in \mathcal{T}({\bm{\theta}_m})}\left\{
\mathcal{L}(\bm{\theta}_m)+ \langle \Delta\bm{\theta}^{\text T},\nabla_{\bm{\theta}}\mathcal{L}(\bm{\theta}_m)\rangle+ \dfrac{1}{2} \Delta\bm{\theta}^\text{T}\mathbf{B}\Delta\bm{\theta}\right\}.
\end{align*}

To aid understanding, the notion of the tangent space associated with a point $\bm{\theta}_m$  in the parameter space corresponds to the set of all possible directions that can be traversed from  $\bm{\theta}$ that yields different directional derivatives w.r.t. the loss function ${\mathcal{L}}$, a map from the parameter space to the real line. Thus, the $\Delta\bm{\theta}$ that are often probed in the optimisation are in fact  members of  $\mathcal{T}({\bm{\theta}_m})$.  
In the above equation, $\langle \Delta\bm{\theta}^{\text T},\nabla_{\bm{\theta}}\mathcal{L}(\bm{\theta}_m)\rangle$ corresponds to an inner product between vectors in  $\mathcal{T}({\bm{\theta}_m})$ where the inner product can be generalised to any Riemannian metric \citep{deFelice:1990hu}.
Replacing the standard inner product with the positive definite $\mathbf{F}^{-1}$ leads\footnote{The derivation holds for any positive scalar multiple of the Fisher. }  to the following optimisation problem in $\mathcal{T}({\bm{\theta}_m})$:
\begin{align}
\label{NGHF:eqn}
\argmin_{\Delta\bm{\theta} \in \mathcal{T}({\bm{\theta}_m}) }
\left\{\mathcal{L}(\bm{\theta}_m)+  \Delta\bm{\theta}^{\text T} \mathbf{F}^{-1}\nabla_{\bm{\theta}}\mathcal{L}(\bm{\theta}_m)+ \dfrac{1}{2} \Delta\bm{\theta}^\text{T}\mathbf{B}\Delta\bm{\theta}\right\}
\end{align}
where each entry of the matrix $\mathbf{F}^{-1}$ is a smooth function of the current estimate $\bm{\theta}_m$.  Differentiating  Eqn. \eqref{NGHF:eqn} and equating to zero leads to following solution:
\begin{align}
\mathbf{B}\Delta\bm{\theta}=-\mathbf{F}^{-1}\nabla_{\bm{\theta}}\mathcal{L}(\bm{\theta}).
\label{eq:NGHFdir}
\end{align}
The right hand side of this above equation corresponds to the NG direction. Hence Eqn.~\eqref{NGHF:eqn} presents a procedure to regulate the scaling and rotation applied by $\mathbf{F}^{-1}$ with a matrix $\mathbf{B}$ chosen in an appropriately understood sense.

\subsection{NGHF: Combining NG and HF based on CG}
\label{ssec:NGHF}

The choice of appropriate $\mathbf{B}$ varies from task to task. In \citep{haider2018}, the prevalence of over-fitting due to criterion mismatch was observed to be highly correlated with the increased sharpness of the NN frame posteriors. In  Sec.~\ref{ssec:GNscaling}, it is discussed how scaling with the inverse of the GN matrix  regulates sharp changes in the entropy of DNN frame posteriors.  Using this insight, in this work $\mathbf{G}$ is employed to regulate the NG descent direction.
Computing the individual inverse matrix scalings in Eqn.~\eqref{NGHF:eqn} directly is  expensive  in terms of both computation and storage. Using the HF approach, each individual matrix scaling is approximated by solving equivalent linear systems using CG.  In Sec.~\ref{sec:HFOpt}  it is shown how the update direction proposed at each CG iteration corresponds to:
\begin{align*}
\Delta \bm{\theta}_{k+1} \gets \Delta \bm{\theta}_{k} + a_{k}\bm{v}_{k}
\end{align*}
where $\bm{v}_{k}$ represents the current conjugate direction. At the first iteration of CG, this is the direction that the  algorithm has been initialised with. In contrast to NG and HF,  the initial direction now corresponds to approximation of the NG direction instead of the gradient. Thus when Eqn.~\eqref{NGHF:eqn} is solved with CG, the resultant update corresponds to:
\begin{align}
   \Delta \bm{\theta} = w_1 \Delta \bm{\theta}_{\text{NG}} + w_2 \Delta \bm{\theta}_{\text{HF}}
\end{align}
which is a weighted  combination of the NG direction and conjugate directions computed using local curvature information. Hence, in this sense we denote this approach NGHF.

\section{Experimental Setup} 
\label{sec:expsetup}
The proposed optimisation framework was  evaluated  on data from the multi-genre broadcast (MGB) challenge \citep{Bell2015} which uses data from a wide range of BBC television programmes, and the effectiveness of the techniques for discriminative sequence training with the MPE loss was found. All systems were trained using a 200 hour training set. The official development set dev.full was split into two subsets. One split corresponds to the official MGB subset, \textbf{dev.sub}, with 5.5 hours data, which is used as the validation set to choose the hyper-parameters and select the best parameter update $\Delta\bm{\theta}$ in Alg. \ref{CGalgo}. To evaluate the generalisation ability to unseen data, an evaluation test set \textbf{dev.sub2} was also created, which consists of 23 hours data from the remaining 35 episodes in the dev.full set. Further detail related to the data preparation  can be found in \citep{Woodland2015}. The input to all models were 40-dim log-Mel filter bank features extended with their delta coefficients, which were normalised at the utterance-level for mean and at the show-level for variance  \citep{Woodland2015}. 
All experiments were conducted using HTK version 3.5 and the PyHTK pipelines\citep{htk,zhang2019pyhtk}.

The RNN models used in the experiments consist of two 1000-dim recurrent layers followed by a 1000-dim feedforward layer. Each recurrent layer is unfolded for 20 steps (from $+$5 to $-$14). Apart from replacing the standard RNN layer with the LSTM layers of the same size, the LSTMs have the same structure as the RNNs.  
The TDNNs used the same structure as in \citep{DanTDNN} with five 1000-dim hidden layers, whose context shifts used to splice the features are $\{-2,-1,0,1,2\}$, $\{-1,2\}$, $\{-3,3\}$, $\{-7,2\}$ and $\{0\}$ from the input to the output layers respectively. 
For all models, the output layer consists of about six thousand output units, with each output corresponding to a context-dependent triphone state obtained by conventional decision tree tying approach. 

The large-batch-based methods, HF, NG, and NGHF, are compared with the mini-batch-based methods, SGD and Adam. For this the LSTM-HMMs, RNN-HMMs, and TDNN-HMMs are trained at the sequence-level with these different optimisers based on the MPE loss. Prior to sequence training, they are trained with SGD on a frame-level CE loss.
To monitor the occurrence of over-fitting and avoid the mismatch between the WER and MPE loss, decoding was performed on both of the validation and evaluation sets,  using a 158k word vocabulary trigram language model.

Since standard SGD and Adam suffer from high data transmission costs in synchronous distributed processing, such optimisers used single GPU training and standard configurations \citep{Su2013,vesely,zhang:2015HTK}. The hyper-parameters associated with SGD and Adam were chosen using grid search such that the improvements obtained from MPE training are closely correlated with WER reductions on the validation set.

Following \citep{kingsbury2012}, all experiments on HF, NG, and NGHF that used the proposed CG-based optimisation framework were performed in a synchronous distributed setting where the gradients were computed across multiple workers in parallel and then accumulated. To achieve a good balance between the reduced training time cost through parallelisation and keeping the data transmission cost low, the gradient batch used in this work contains 25 hours of data. When using four workers instead of one worker to collect the gradients over a 25 hour gradient batch, the time cost is reduced from 150 minutes to 37 minutes, which is an almost a linear reduction in the time cost by a factor of the number of workers. The method itself allows full parallelisation of the gradient calculation stage such that one worker is used per utterance. Parallelisation is not only constrained by the number of workers available, but also the additional overhead time-cost due to further parallelisation and data transmission\footnote{The optimal degree of parallelisation will vary depending on the computing infrastructure used.}.
To reduce the time cost associated with the individual CG iterations, a 0.5 hour CG batch was used throughout the experiments, whose data were uniformly sampled from the entire training set.
It should be noted that one worker was used for CG, although in theory the maximum number of workers allowed for CG could be the same as the number of utterances in the CG batch. 
It was found that running CG for 5-8 iterations was sufficient to find an parameter update that could yield a reasonable reduction in the loss value. In terms of computational cost, the CG worker was found to take about 30 minutes to execute 8 CG iterations for LSTM-HMMs. 
All timing information was obtained on a machine with four Tesla P100 GPUs and a 14 core Intel Xeon CPU E5-2680 v4 at 2.40GHz.

From Table \ref{tab:timecost}, it is clear that validating the performance at each CG iteration takes the largest proportion of the overall time cost. Recalling that the goal of CG is to minimise a quadratic approximation of the loss function, 
through the validation of the update obtained by each CG iteration on the CG batch, the validation stage checks whether CG follows a path in the parameter space where the quadratic approximation still holds. 
Although a validation stage was performed after every CG iteration in all the  experiments in this paper, we empirically found that the check can be performed less frequently to reduce the time cost.
\begin{table}[h]
\centering
\caption{The proportion of time cost for running the CG algorithm for NGHF for 8 iterations using a single NVidia P100 GPU and a single core of Intel Xeon CPU E5-2680 v4 at 2.40GHz. The time cost for loading the lattices into the memory is excluded from the calculation since it can vary considerably depending on the lattice implementation and the speed of the storage system.}
\begin{tabular}{lc}
\toprule
Procedure &  \%Time cost \\
\midrule
Modified forward propagation & 15.1 \\
EBP & 7.8 \\
Collecting statistics over lattices & 4.1 \\
Evaluating the performance of each $\Delta\bm{\theta}$  & 73.0 \\
\bottomrule
\end{tabular}
\label{tab:timecost}
\end{table}


\section{Experimental Results}
\label{sec:ExpResults}

This section first compares the various optimisers using LSTM models (Sec.~\ref{sec:LSTMResults}). The evolution of the MPE loss function during discriminative sequence training is discussed, as well as the final WER on the evaluation set. The performance when using both sigmoid and ReLU activation functions with RNN and TDNN models is then investigated for the different optimisation approaches (Sec.~\ref{sec:ActFnResults}).

\begin{figure}
\centering
\includegraphics[width=\linewidth]{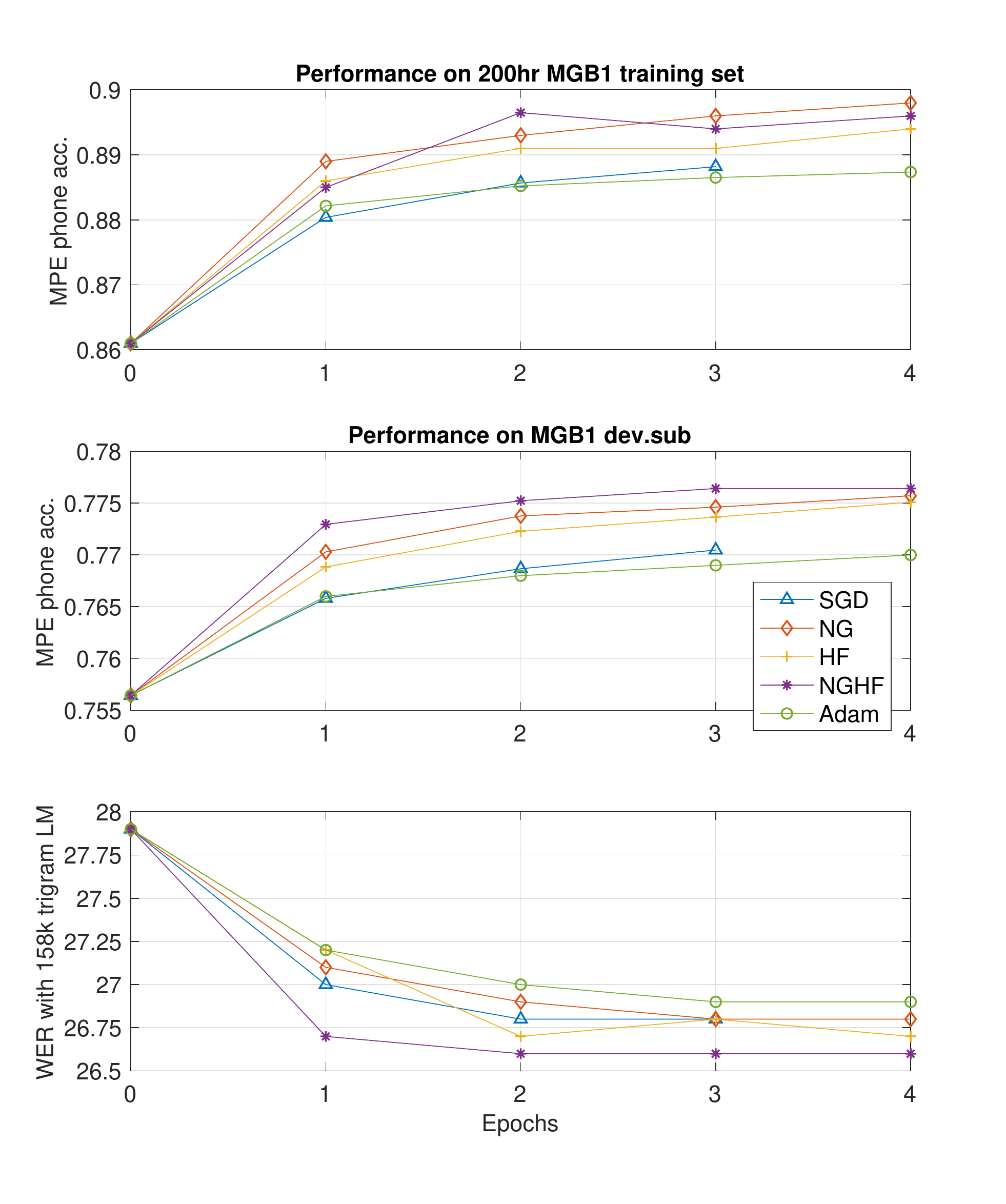}
\caption{The evolution of the performance of the LSTM-HMMs with different optimisers. The first two plots show the phone accuracy on a subset of the training set and the validation set dev.sub. 
The third plot shows the absolute reductions in \%WERs  on the validation set.}
\label{plot}
\end{figure}

\begin{table}[h]
\centering
\caption{LSTM-HMM performance on the validation set with different optimisers. ``Best epoch'' gives the epoch model with the best performance. ``\#Update'' shows the number of updates used in the MPE training, and ``k'' stands for one thousand. ``MPE Acc.'' gives the MPE accuracy (negative of MPE loss value). Although NG, HF, and NGHF used much fewer updates, the numbers of utterances processed in each epoch are similar to those of SGD and Adam.}
\label{tab:gen1}
\begin{tabular}{lcccc}
\toprule
Optimiser & Best epoch & \#Update & MPE Acc & \%WER \\
\midrule
CE  & --- & --- & 0.765 & 27.9 \\
\midrule
SGD & 3 & 420k & 0.771 & 26.8 \\
Adam & 4 & 560k & 0.770 & 26.9 \\
NG & 3 & 24 & 0.775 & 26.8 \\
HF & 2 & 16 & 0.772 & 26.7 \\
NGHF & 2 & 16 & 0.775 & \textbf{26.6} \\
\bottomrule
\end{tabular}
\end{table}


\subsection{LSTM-HMMs Results} \label{sec:LSTMResults}
Figure~\ref{plot} shows how MPE phone accuracy (negative of the MPE loss) evolves during training for the LSTM-HMMs with different optimisers, and the corresponding changes in WER. Table~\ref{tab:gen1} summarises the results on the validation set for the best epoch. From Fig.~\ref{plot}, NGHF is the most effective method in terms of both improving the MPE accuracy and reducing the WER. 
By combining the KL-divergence with the local curvature information obtained from the Hessian matrix, the  NGHF method achieves a greater WER reduction with just 8 parameter updates (\textit{i.e.} one epoch) than the converged model obtained trained using SGD or Adam using hundreds of thousands of updates. 
Table~\ref{tab:gen1}, shows that MPE training using NGHF results in a relative 4.7\% WER reduction (WERR) over the CE trained model. 
It can also be seen that all of the three CG-based second-order optimisers require far fewer updates and fewer training epochs to converge to a good solution than SGD and Adam.

To ensure the performance difference of LSTM-HMMs trained with different optimisers can be generalised to unseen data, all models were tested on the evaluation set, which was not use for setting hyper-parameters.  
Table \ref{tab:gen2} summarises the results of the LSTM-HMMs on the evaluation set. From the results in Table \ref{tab:gen2}, 
MPE with NGHF achieves the lowest WERs among all five optimisers, and obtains a 3.4\% relative WERR compared to the CE trained model. Compared to both SGD and Adam, NGHF achieves a 1\%  larger WERR while requiring far fewer parameter updates and also fewer training epochs. 
This improvement was found to be statistically significant at the 0.1\% level for dev.sub2\footnote{Statistical significance tests in this paper use a sign test on WER differences at the episode level for the 35 programme episodes included in dev.sub2.}.

\begin{table}[h]
\centering
\caption{\%WERs for MPE trained LSTM-HMMs on the evaluation set with different optimisers. ``CE'' refers to the SGD trained CE system, while the others refer to the MPE systems trained with the corresponding optimisers. }
\label{tab:gen2}
\begin{tabular}{cccccc}
\toprule
CE & SGD & Adam & NG & HF & NGHF \\
\midrule
29.3 & 28.6 & 28.6 & 28.6 & 28.4 & \textbf{28.3} \\
\bottomrule
\end{tabular}
\end{table}

\subsection{TDNN-HMMs and RNN-HMMs with sigmoid and ReLU Activation Functions} \label{sec:ActFnResults}
This section presents a comparison between 
RNNs and TDNNs with both ReLU and sigmoid activation functions when using  the various optimisers for MPE training. It is well-known that ReLU and sigmoid models require very different configurations in SGD-based training: ReLU models often need a learning rate and the range for random initialisation a factor of 4 to 8 times smaller than those used for sigmoid models, and that  discriminative sequence training with ReLU activation functions often results in over-fitting to MBR loss functions\footnote{By over-fitting we here mean a low correlation between reduction in MBR loss and reduction in WER.}. We claim that such a difference still exists with the second-order optimisers since the second-order derivatives of ReLU and sigmoid are very different\footnote{For ReLU $h''(a_{t,j})=0$ and for sigmoid  $\sigma''(a_{t,j})=\sigma(a_{t,j})(1-\sigma(a_{t,j}))(1-2\sigma(a_{t,j}))$}, which causes a large difference in the calculation of directional derivatives \citep{BishopPRMF}.  


Table~\ref{tab:gen4} gives the number of updates required by each type of optimiser to find a good  solution and Table~\ref{tab:gen5} compares  the efficacy of the trained models with different optimisers on the evaluation set.  For the sigmoid models, using NGHF obtains the largest reductions in WER from MPE training, yielding a 5\% WERR for RNN and 6\% WERR for TDNN, over the  relevant CE trained models. Compared to SGD and Adam, NGHF achieves a WERR of 3.5\% for sigmoid RNN and 1\% for sigmoid TDNN while using far fewer updates. These WER reductions were found to be statistically significant (p < 0.001) on dev.sub2.
 
Regarding the ReLU models, it was found to be very difficult to get reasonable WER reductions using SGD, Adam and NG. With NG in particular, sequence training was found to suffer from over-fitting  due to the mismatch between the MPE loss and the WERs from the very beginning of the MPE training. On the validation set, NG was observed to achieve excellent generalisation performance in terms of MPE loss but such improvements failed  to correlate well with the WER reductions. For the TDNN ReLU model, MPE training with NGHF gives the the biggest WER reduction, which achieves a WERR of 4.2\% compared to the CE trained model. In contrast to SGD and Adam, NGHF achieves WERRs of 2\%. For the ReLU RNN model,  although HF produced the lowest WER, its WERR over the NGHF method was not found to be statistically significant (at the 5\% level). 

\begin{table}[h]
\centering
\caption{Number of MPE updates required by different optimisers (``k'' stands for 1000) for ReLU and sigmoid ($\sigma$) RNN and TDNN models. Although NG, HF, and NGHF used far fewer updates, the numbers of utterances processed in each epoch are similar to those of SGD and Adam. }
\label{tab:gen4}
\begin{tabular}{lccccc}
\toprule
Model & SGD & Adam & NG & HF & NGHF \\
\midrule
ReLU RNN & 100k & 100k & 0 & 24 & 16 \\
$\sigma$ RNN & 440k & 440k & 40 & 40 & 40 \\
ReLU TDNN & 330k & 100k & 0 & 32 & 32 \\
$\sigma$ TDNN & 440k & 440k & 48 & 48 & 40 \\
\bottomrule
\end{tabular}
\end{table}

\begin{table}[h]
\centering
\caption{\%WER for ReLU and sigmoid ($\sigma$) RNN and TDNN models on the evaluation set with different optimisers. ``CE'' refers to the SGD trained CE system, while other columns refer to the MPE systems trained with the corresponding optimisers. }
\label{tab:gen5}
\begin{tabular}{lcccccc}
\toprule
Model & CE & SGD & Adam & NG & HF & NGHF \\
\midrule
ReLU RNN & 30.3 & 30.3 & 30.2 & 30.3 & \textbf{29.6} & 29.7 \\
$\sigma$ RNN & 32.2 & 31.6 & 31.6 & 30.6 & 30.7 & \textbf{30.5} \\
ReLU TDNN & 30.6 & 29.8 & 29.8 & 30.6 & 29.6 & \textbf{29.3} \\
$\sigma$ TDNN & 29.9 & 28.5 & 28.3 & 28.2 & 28.5 & \textbf{28.1} \\
\bottomrule
\end{tabular}
\end{table}

\section{Discussion}
\label{sec:Discussion}
The optimisation framework presented in this work provides a general method to flexibly train NN models with any arbitrary smooth loss using a large batch NG descent or second-order method in a data-parallel manner. Although the work has  applied the framework in a centralised distributed training environment, the framework presented can also be applied in a decentralised distributed training environment as well \citep{zhangw2019a,zhangw2019b}. 

Compared to the traditional HF approach, the novel modifications proposed in this paper overcome the computational drawbacks of requiring hundreds of CG iterations that have been cited as a key issue with HF \citep{Martens2015}. The experimental results presented with the LSTM, TDNN and RNN models show how effective updates from only a small number of CG iterations can be obtained, and this approach to model training leads  to significant WER reductions.  
The efficacy of the novel preconditioning approach mentioned in Sec. \ref{sec:CGpredcond} can be clearly seen when training  the LSTM model. Table \ref{tab:gen2} shows that NG, HF and NGHF 
all yield greater WER reductions in comparison  to the stochastic-gradient-based approaches while using far fewer updates.  
Furthermore, this paper presents a novel procedure to regulate NG learning whose efficacy can be seen in Table \ref{tab:gen4} and \ref{tab:gen5}.  For the ReLU-based TDNN and RNN models, the regularised NG approach (\textit{i.e.} NGHF) is more adept in following a path in the parameter space where minimising w.r.t. the training criterion correlates with WER reductions.
This can be very important when applying NG to other structures and non-ASR tasks, since ReLU is widely used by the models of computer vision \citep{Simonyan2015} and natural language processing \citep{Vaswani2017}.

In addition to the differences in WER and the number of updates reported in Sec.~\ref{sec:ExpResults}, 
 each of the optimisers require different amounts of computation and GPU memory. 
For SGD, the input and output values of each layer are calculated in forward-propagation. In the backward-propagation pass, the derivatives w.r.t. the input/output values, as well as the gradients of the parameters are calculated.
The parameters $\bm{\theta}$, gradients $\nabla_{\bm{\theta}}\mathcal{L}(\bm{\theta})$, and update values $\Delta\bm{\theta}$  each require the same amount of space when they are stored in GPU memory. The input/output values and their derivatives are also stored in GPU memory, and the required storage  depends on the number of frames in the mini-batch. 
For Adam, extra calculation and GPU memory are required to compute and store $(\nabla_{\bm{\theta}}\mathcal{L}(\bm{\theta}))^2$ and the relevant second raw moment estimate for every frame in the training set \citep{kingma2014adam}. 

NG, HF, and NGHF all have the same computation and storage complexity as SGD for the gradient accumulation stage.  In addition to the standard forward-propagation and backward-propagation procedures, each CG iteration uses an extra modified forward-propagation procedure to calculate the directional derivatives w.r.t. input/output values, as described in Sec. \ref{ssec:mbr}. This results in more memory usage than Adam. The extra computation cost in the CG stage applies to the CG batch, which is only a small portion of the training set. Note that the number of CG iterations used in NG, HF, and NGHF differs in our experiments (see Sec.~\ref{sec:expsetup}). An extra validation stage is performed after every CG iteration, which increases the computation cost of the proposed methods although it can be performed less frequently with a smaller amount of data. 
Despite the increased cost, the computation within each gradient batch or CG batch can be easily parallelised (as shown in Fig.~\ref{fig:framework}) which makes it much easier to use many workers with no further approximations than for SGD or Adam. When applied to a much larger training set, it may be unnecessary to scale up the size of the CG batch, which would reduce the relative increase in computation cost.

\section{Conclusions}
A CG-based synchronous distributed optimisation framework for discriminative sequence training has been proposed in this paper, which has the flexibility to combine NG with a second-order optimisation method, such as HF. 
This framework has the same advantages as HF in providing stable loss value reductions and inherently suitable for parallel computing, yet also has improved numerical issues in CG and improved performance for models with shared parameters. 
The framework was evaluated using ASR experiments with the training and test data from the MGB challenge. It was shown that the improved CG method can find effective parameter updates resulting in a better improvement in MPE loss often with far fewer CG iterations. Furthermore, when applied in a setting where NG is combined with HF, the resulting NGHF method generates models with better generalisation ability with far fewer parameter updates when compared with SGD and Adam. The method also can be efficiently parallelised across multiple GPUs without making any approximations. 

\section*{Acknowledgement}
The authors are grateful to the anonymous reviewers for their valuable suggestions that helped improve an earlier version of this paper.

\printcredits

\bibliographystyle{bibstyle_arxiv}

\bibliography{refs_arxiv}





\end{document}